\documentclass[10pt,twocolumn,letterpaper]{article}

\usepackage[pagenumbers]{cvpr}

\definecolor{linkblue}{rgb}{0.21,0.49,0.74}
\usepackage[breaklinks,colorlinks,allcolors=linkblue]{hyperref}
\hypersetup{
    hypertexnames=false,
    pdftitle={LLMVC: Large Language Model-Guided Generative Video Compression},
    pdfauthor={Lebin Zhou, Zhenghao Chen, Nam Ling, Wei Wang, Wei Jiang}
}

\title{LLMVC: Large Language Model-Guided Generative Video Compression}

\author{%
Lebin Zhou\textsuperscript{1}\quad
Zhenghao Chen\textsuperscript{2}\thanks{Corresponding authors.}\quad
Nam Ling\textsuperscript{1}\quad
Wei Wang\textsuperscript{3}\quad
Wei Jiang\textsuperscript{3}\footnotemark[2]\\[4pt]
{\normalsize\textsuperscript{1}Santa Clara University, Santa Clara, CA, USA}\\
{\normalsize\textsuperscript{2}The University of Newcastle, Callaghan, NSW, Australia}\\
{\normalsize\textsuperscript{3}Futurewei Technologies, Inc., San Jose, CA, USA}\\[4pt]
{\small\texttt{\{lzhou,nling\}@scu.edu}}\\
{\small\texttt{zhenghao.chen@newcastle.edu.au}}\\
{\small\texttt{\{rickweiwang,wjiang\}@futurewei.com}}
}

\begin{document}
\setcounter{footnote}{1} 
\maketitle

\begin{abstract}

  Recent generative video methods achieve strong perceptual compression by encoding videos into compact latent representations and reconstructing them with powerful generative decoders. However, visual encoders must compress both high-level semantics and dense visual details into a unified latent space, which complicates entropy coding and decoding and may cause semantic drift under aggressive compression. In this work, we propose Large Language Model-Guided Video Compression (LLMVC), which leverages the summarization capability of frozen video large language models (Video-LLMs) to generate compact semantic tokens aligned with temporal latent slices for guiding video compression. Specifically, Hybrid Bidirectional Bimodal Attention (HBBA) enables bidirectional interactions between visual and semantic features during encoding and decoding. Semantic-Conditioned Cross-Attention (SCCA) incorporates semantic features into probability modeling during entropy coding. These semantic features provide high-level guidance throughout the pipeline, improving representation quality and encouraging semantic consistency through a transmitted text side channel. Experiments on UVG, MCL-JCV, and HEVC show that LLMVC consistently improves compression performance over traditional and recent neural video compression methods in terms of LPIPS and DISTS, while also achieving better fidelity than recent perceptual codecs.

\end{abstract}

\section{Introduction}
\label{sec:intro}

With video data accounting for a dominant share of internet traffic, effective video compression has become increasingly important. Traditional codecs~\cite{vvc,hevc,avc} and neural video codecs~\cite{fvc,dcvc-rt,dcvc,dcvc-hem,dcvc-dc,dcvc-fm,dvc,dcvc-tcm,hlvc,rlvc,plvc} are largely designed to minimize generic distortion, while recent research has increasingly emphasized compression methods that better align with human perceptual quality. Recent studies~\cite{li2024extreme,DiffVC,glc} leverage video generation models (VGMs) to encode videos into compact latent representations and reconstruct them with powerful VGM decoders. These approaches achieve promising perceptual reconstruction quality, particularly at low bitrates, by capturing the overall visual structure of a video in a compressed latent space.

However, generative video compression methods typically represent all information from the raw video signal in a unified latent space. This latent must capture both high-level semantic content, such as scenes, objects, and actions, and low-level visual details, such as textures and colors. In most existing approaches, these two types of information are implicitly entangled, which can reduce efficiency during entropy coding and decoding. Semantic consistency must be preserved through visual features alone, producing dense and less factorized latents, while the decoder relies on the same latent to recover both semantic structure and fine-grained details. Consequently, higher-fidelity reconstruction often requires substantially more bits for only marginal improvements in perceptual quality.

Existing evidence suggests that perceptual quality depends mainly on high-level semantic consistency, while low-level details play a secondary role~\cite{chen2024topiq,fu2023dreamsim}. This raises a fundamental question: \textit{should the latent encode all information uniformly, or can compression benefit from explicitly disentangling semantic content from low-level appearance?} Video Large Language Models (Video-LLMs) summarize complex scenes into concise textual descriptions~\cite{llava1.5,nvila}, providing semantic-only representations that are substantially more compact than dense visual features. However, they omit low-level details and cannot directly reconstruct the original visual signal. This leads to a second question: \textit{can a comprehensive latent representation be aligned with semantic-only features to strengthen high-level semantic structure, while retaining only the minimal low-level visual information required for efficient entropy coding and high-quality decoding?}

To answer these questions, we propose \textbf{LLM-Guided Generative Video Compression (LLMVC)}, a text-guided video compression framework that leverages extremely compact semantic features (i.e., text tokens) to guide video compression. Specifically, we introduce a \textbf{Semantic-Guidance Mechanism} that consistently aligns VGM latent representations with semantic-only features generated by a Video-LLM for the corresponding temporal latent slices during encoding, entropy coding, and decoding. To enable this alignment, we use Hybrid Bidirectional Bimodal Attention (HBBA) during encoding and decoding and Semantic-Conditioned Cross-Attention (SCCA) during entropy coding to provide semantic-guidance mechanisms for latent feature modeling. Based on these mechanisms, we develop three semantic-guidance modules: Semantic Latent Alignment (SLA) for encoding, Semantic-Conditioned Entropy Model (SCEM) for entropy coding, and Semantic Texture Enhancement (STE) for decoding.

These modules progressively refine the latent representation throughout the codec, producing more structured representations that better preserve high-level semantics while retaining the low-level information needed for efficient entropy coding and faithful reconstruction. The descriptions are losslessly transmitted as side information, and their bitrate is included in the total rate. Experiments on UVG, MCL-JCV, and HEVC show that the proposed LLMVC consistently improves perceptual reconstruction quality over both traditional codecs and recent neural video compression methods.

Our contributions can be summarized as follows:
\begin{itemize}
    \item We propose an LLM-guided generative video compression framework that introduces a compact semantic channel aligned with temporal latent slices, using Video-LLM-generated text tokens to guide representation learning in a VGM-based latent codec.
    \item  We develop three semantic-guidance modules using Hybrid Bidirectional Bimodal Attention (HBBA) and Semantic-Conditioned Cross-Attention (SCCA) at different stages of the rate-distortion pipeline. Semantic Latent Alignment (SLA) and Semantic Texture Enhancement (STE) improve reconstruction quality (distortion), while Semantic-Conditioned Entropy Model (SCEM) improves probability modeling for entropy coding (rate).
    \item We evaluate LLMVC on standard benchmarks including UVG, MCL-JCV, and HEVC, and demonstrate consistent improvements in perceptual quality over traditional codecs and recent neural video compression methods in terms of LPIPS and DISTS.
\end{itemize}

\section{Related Work}
\label{sec:relatedworks}

\subsection{Neural Video Compression}
Neural video compression began with optical-flow-based residual coding~\cite{fvc, dvc, hlvc, rlvc}. Contextual coding methods~\cite{st-xct, dcvc, dcvc-hem, dcvc-dc, dcvc-fm, dcvc-tcm} later replaced explicit residuals with learned feature priors, while GAN-based approaches~\cite{gvc, cgvct, hvfvc, plvc} introduced adversarial training for sharper textures. Recently, tokenizer-based frameworks~\cite{glc, tvc} compress videos into compact latent representations, but still rely on implicit visual features without explicit semantic-guidance mechanisms. Generative codecs further improve perceptual quality: GLC~\cite{glc} and DiffVC~\cite{DiffVC} adapt pre-trained image generative models through frame-wise generative embeddings or diffusion-based enhancement, but their image-domain priors lack explicit temporal modeling and may cause flickering or motion inconsistency. GNVC-VD~\cite{gnvcvd} instead jointly encodes spatio-temporal latents and refines them with a video-native diffusion transformer for temporally coherent reconstruction. Nevertheless, these methods remain purely visual and leave high-level semantics implicit. Our work addresses this limitation using semantic features derived from Video-LLM-generated descriptions to guide encoding, entropy coding, and decoding.

\subsection{Text-Driven Visual Compression}
Text-driven visual compression uses natural language as a compact semantic bottleneck. Early methods employed text-guided diffusion models to synthesize details at ultra-low bitrates~\cite{perco_2023,textsketch_2023}, but incurred high latency. Later approaches integrated text directly into codec features: TGIC~\cite{tgic_2023} conditions both the encoder and decoder on text, while TACO~\cite{taco_2024} injects CLIP embeddings into the encoder for semantic alignment. Recent methods further use large multimodal models (LMMs) to generate descriptions for diffusion-based decoding~\cite{misc_2024} or interactively optimize coding configurations~\cite{compx_2025}. However, many prior approaches rely on static or manually provided text and capture only coarse semantics, while recent LMM-based methods still apply semantic-guidance mechanisms to limited codec stages without jointly guiding encoding, entropy coding, and decoding. Our work instead uses a Video-LLM to automatically generate compact semantic tokens aligned with temporal latent slices and propagate them throughout the compression pipeline.

\section{Methodology} 

\begin{figure*}[!t]
  \centering
  \includegraphics[width=1.0\linewidth]{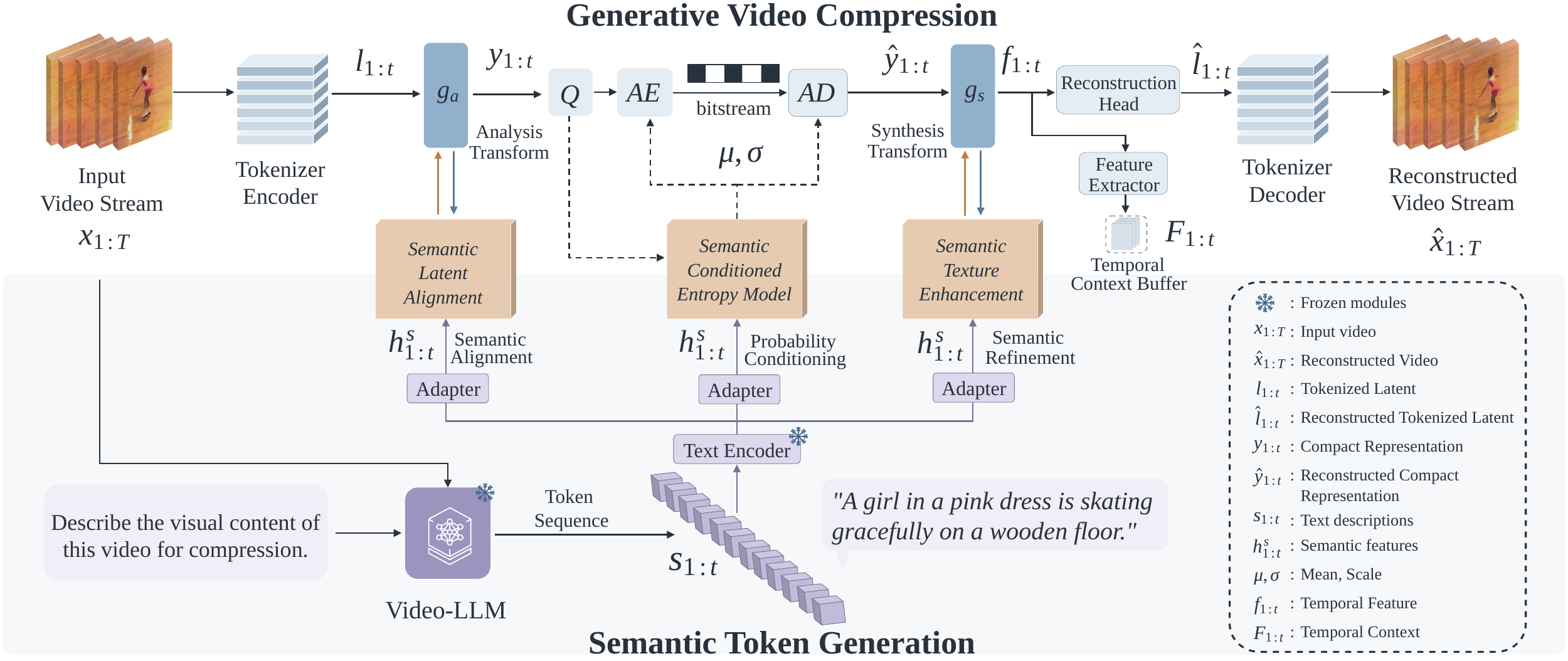}
  \caption{Overview of the proposed LLMVC framework. A tokenizer encoder maps the input video to temporal latents, while a Video-LLM generates corresponding descriptions. A frozen text encoder and trainable adapters map them to semantic features. Semantic Latent Alignment (SLA) aligns intermediate visual features from the analysis transform $g_a$ with semantic features, Semantic-Conditioned Entropy Model (SCEM) uses semantic features to guide probability estimation, and Semantic Texture Enhancement (STE) refines intermediate decoded features from the synthesis transform $g_s$. The tokenizer decoder reconstructs the video from the decoded latents. Q denotes quantization, and AE and AD denote arithmetic encoding and decoding. The legend on the right defines the framework symbols.}
  \label{fig:llmvc}
\end{figure*}

\subsection{Preliminary}

Existing methods typically introduce high-level priors at individual codec stages, such as latent representation learning during encoding~\cite{movie_2026} or entropy modeling~\cite{llmc2, llmc3}. In contrast, LLMVC uses semantic-guidance modules across multiple stages to reduce bitrate and improve reconstruction quality, as illustrated in Fig.~\ref{fig:llmvc}.

A frozen Video-LLM generates descriptions $s_{1:t}$ aligned with the temporal latent slices, where $t$ is the number of temporal latents. We convert these descriptions into semantic features $h_{1:t}^s$ and integrate them through Hybrid Bidirectional Bimodal Attention (HBBA) during encoding and decoding and Semantic-Conditioned Cross-Attention (SCCA) during entropy coding. Based on these mechanisms, we introduce three semantic-guidance modules.

\begin{itemize}

\item \textbf{Semantic Latent Alignment (SLA)} aligns intermediate visual features from the analysis transform with semantic features, encouraging more compact and semantically structured representations (\textit{distortion side}).

\item \textbf{Semantic-Conditioned Entropy Model (SCEM)} conditions probability estimation on semantic features, providing additional context to improve entropy coding efficiency (\textit{rate side}).

\item \textbf{Semantic Texture Enhancement (STE)} refines decoded representations using semantic-guidance mechanisms, enhancing fine-grained textures after quantization (\textit{distortion side}).
\end{itemize}

\begin{figure*}[!t]
  \centering
  \includegraphics[width=1.0\linewidth]{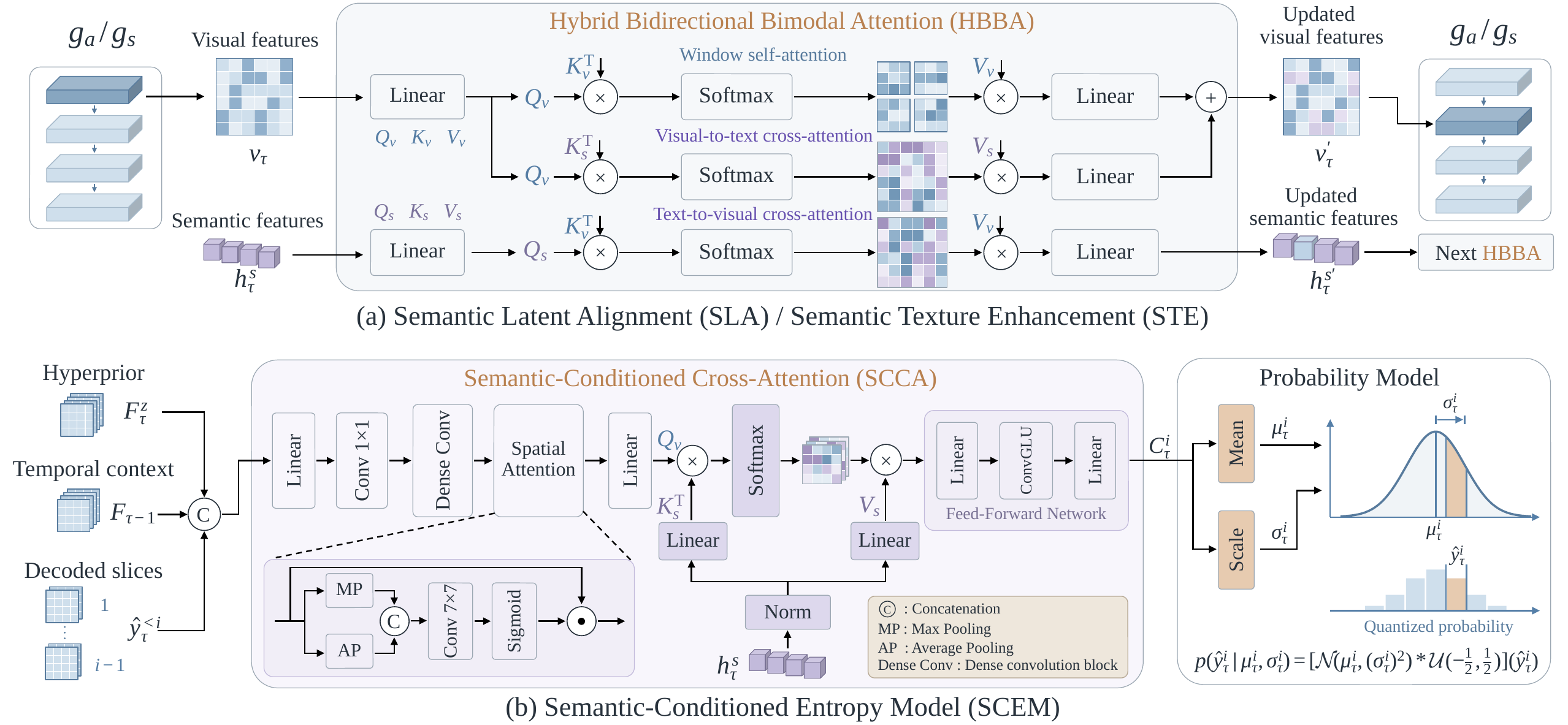}
\caption{Semantic-guidance modules in LLMVC. (a) SLA and STE use Hybrid Bidirectional Bimodal Attention (HBBA) to exchange information between semantic features and intermediate visual features from the analysis transform $g_a$ and synthesis transform $g_s$, producing updated features $v'_\tau$ and $h_\tau^{s\prime}$. (b) SCEM uses Semantic-Conditioned Cross-Attention (SCCA) to combine hyperprior features $F_\tau^z$, temporal context $F_{\tau-1}$, and previously decoded slices $\hat{y}_\tau^{<i}$ with semantic features $h_\tau^s$, producing context $C_\tau^i$ for predicting the Gaussian mean $\mu_\tau^i$ and scale $\sigma_\tau^i$ for entropy coding. Here, $\tau$ indexes temporal latents and $i$ indexes channel slices. Q, K, and V denote queries, keys, and values, respectively. The subscripts $v$ and $s$ indicate visual and semantic features.}
  \label{fig:codec_scca}
\end{figure*}

\subsection{Semantic-Guidance Mechanism}
\label{sec:textual}

\noindent\textbf{Semantic Feature Extraction.} The description $s_\tau$ summarizes the frame group associated with temporal latent $l_\tau$, where $\tau\in\{1,\ldots,t\}$ indexes the temporal latents. The first description corresponds to the first video frame, while each subsequent description corresponds to the next $r$ frames, where $r$ denotes the tokenizer's temporal downsampling factor. Denote these index sets by $\mathcal{G}_1=\{1\}$ and $\mathcal{G}_\tau=\{2+(\tau-2)r,\ldots,1+(\tau-1)r\}$ for $\tau\geq2$. A frozen text encoder followed by a trainable adapter $\phi$ converts each description into semantic features given by
\begin{equation}
\label{eq:semantic_features}
h_\tau^s=\phi\bigl(\mathrm{TextEnc}(s_\tau)\bigr).
\end{equation}
Here, $h_\tau^s\in\mathbb{R}^{L_\tau\times d_s}$, where $L_\tau$ is the text sequence length and $d_s$ is the semantic feature dimension.

\noindent\textbf{Hybrid Bidirectional Bimodal Attention (HBBA).}
SLA and STE use HBBA to exchange information between semantic features and intermediate visual features from the analysis transform $g_a$ and synthesis transform $g_s$ (Fig.~\ref{fig:codec_scca}(a)). These interactions alternate with visual processing blocks throughout encoding and decoding. Linear projections produce queries, keys, and values $[Q_v,K_v,V_v]$ from visual features $v_\tau$ and $[Q_s,K_s,V_s]$ from semantic features $h_\tau^s$.

\noindent\textbf{(1) Window self-attention} captures local spatial relationships among visual features within each window, modeling fine-grained structures and textures.

\noindent\textbf{(2) Visual-to-text cross-attention} allows each visual position to attend to all semantic features associated with the current temporal latent.

\noindent\textbf{(3) Text-to-visual cross-attention} allows semantic features to attend to visual features across the entire feature map, integrating information from different windows.

The outputs of window self-attention and visual-to-text cross-attention are combined to update the visual features $v'_\tau$, while text-to-visual cross-attention updates the semantic features $h_\tau^s$ to $h_\tau^{s\prime}$. The updated visual features are passed to the next visual block, while the updated semantic features are carried forward to the next HBBA interaction.

\noindent\textbf{Semantic-Conditioned Cross-Attention (SCCA).}
Within SCEM, SCCA takes the hyperprior features $F_\tau^z$, temporal context $F_{\tau-1}$, and previously decoded slices $\hat{y}_\tau^{<i}$ as visual context, together with the corresponding semantic features $h_\tau^s$ (Fig.~\ref{fig:codec_scca}(b)), where $i$ indexes channel slices. A dense convolution block and spatial attention based on channel-wise average and max pooling process the visual context to form queries $Q_v$. The normalized semantic features provide keys $K_s$ and values $V_s$. Cross-attention retrieves semantic information relevant to the visual context, and a feed-forward network refines the resulting features to produce the context $C_\tau^i$ for probability estimation.

\subsection{LLM-Guided Video Compression Pipeline}
\label{sec:pipeline}

We describe the full compression pipeline in four steps, illustrating how the corresponding semantic features are integrated at each stage (Fig.~\ref{fig:llmvc}).

\noindent\textbf{Step~1. Semantic Description Generation and Coding.} The Video-LLM generates descriptions $s_{1:t}$ aligned with the temporal latents at the encoder. These descriptions are converted into text tokens, whose IDs are losslessly entropy-coded as side information using an adaptive probability model based on symbol frequency statistics. At both ends, a frozen text encoder followed by a trainable adapter converts the tokens of each description $s_\tau$ into the corresponding semantic features $h_\tau^s$.

\noindent\textbf{Step~2. Encoding with Semantic Alignment.}
The tokenizer encoder maps the input video $x_{1:T}$ of $T$ frames to a latent sequence $l_{1:t}$, where $t=1+(T-1)/r$ for an input padded as needed to match the temporal grouping. The latent codec then encodes each temporal latent $l_\tau$ sequentially, using the temporal context $F_{\tau-1}$ derived from previously decoded representations. SLA uses HBBA to align intermediate visual features from $g_a$ with semantic features $h_\tau^s$, encouraging more compact and semantically structured representations (Fig.~\ref{fig:codec_scca}(a)). We denote the overall mapping that alternates $g_a$ blocks with SLA interactions by $g_a^{\mathrm{SLA}}$. The hyperprior analysis transform $h_a$ maps the resulting compact representation $y_\tau$ to a hyperlatent $z_\tau$, which is quantized and entropy-coded. The recovered hyperlatent $\hat{z}_\tau$ is then mapped by the hyperprior synthesis transform $h_s$ to hyperprior features $F_\tau^z$ for probability estimation. These operations are expressed as
\begin{equation}
\label{eq:semantic_encoding}
\begin{gathered}
l_{1:t}=\operatorname{Enc}(x_{1:T}),\qquad y_\tau=g_a^{\mathrm{SLA}}(l_\tau,F_{\tau-1},h_\tau^s),\\
z_\tau=h_a(y_\tau),\qquad \hat{z}_\tau=Q(z_\tau),\qquad F_\tau^z=h_s(\hat{z}_\tau).
\end{gathered}
\end{equation}
Here, $Q$ denotes quantization. The first temporal latent is encoded without temporal context.

\noindent\textbf{Step~3. Entropy Coding with Probability Conditioning.} The compact representation $y_\tau$ is partitioned into $N_s$ channel slices, which are coded sequentially. For slice $i\in\{1,\ldots,N_s\}$, SCEM uses SCCA to combine the visual context with the corresponding semantic features $h_\tau^s$, producing the semantic-conditioned context (Fig.~\ref{fig:codec_scca}(b))
\begin{equation}
\label{eq:semantic_context}
C_\tau^i=\operatorname{SCCA}\!\left([F_\tau^z,F_{\tau-1},\hat{y}_\tau^{<i}],h_\tau^s\right).
\end{equation}
The semantic-conditioned context $C_\tau^i$ is used to predict the parameters of a Gaussian distribution, with mean $\mu_\tau^i$ and scale (standard deviation) $\sigma_\tau^i$. Each slice $y_\tau^i$ is quantized to $\hat{y}_\tau^i$, whose probability model is
\begin{equation}
\label{eq:slice_probability}
p(\hat{y}_\tau^i\mid\mu_\tau^i,\sigma_\tau^i)=\left(\mathcal{N}(\mu_\tau^i,(\sigma_\tau^i)^2)*\mathcal{U}(-\tfrac12,\tfrac12)\right)(\hat{y}_\tau^i).
\end{equation}
This conditioning enables the entropy model to exploit semantic context beyond the visual features. The quantized slices are entropy-coded using the predicted probabilities and concatenated after decoding to form $\hat{y}_\tau$.

\noindent\textbf{Step~4. Decoding with Semantic Refinement.} STE uses HBBA to refine intermediate decoded representations from $g_s$, enhancing fine-grained textures after quantization (Fig.~\ref{fig:codec_scca}(a)). The overall mapping $g_s^{\mathrm{STE}}$ alternates $g_s$ blocks with STE interactions to produce temporal features $f_\tau$. The reconstruction head (RecHead) maps them to $\hat{l}_\tau$, while the feature extractor produces the temporal context $F_\tau$ for subsequent temporal latents. The tokenizer decoder reconstructs the video from the reconstructed latent sequence. These operations are expressed as
\begin{equation}
\label{eq:semantic_decoding}
\begin{gathered}
f_\tau=g_s^{\mathrm{STE}}(\hat{y}_\tau,F_{\tau-1},h_\tau^s),\qquad \hat{l}_\tau=\operatorname{RecHead}(f_\tau),\\
F_\tau=\operatorname{FeatureExtractor}(f_\tau),\qquad \hat{x}_{1:T}=\operatorname{Dec}(\hat{l}_{1:t}).
\end{gathered}
\end{equation}

\subsection{Optimization}
\label{sec:training}

We first optimize the tokenizer and latent codec for rate-distortion performance, then jointly fine-tune them with the semantic-guidance modules.

\noindent\textbf{Stage~1. Rate-Distortion Pretraining.}
In Stage~1, we train the tokenizer and VGM-based latent codec without semantic-guidance modules. The training objective is
\begin{equation}
\label{eq:stage1_objective}
    \mathcal{L}_1
    =
    \lambda \mathcal{D}(x_{1:T}, \hat{x}_{1:T})
    + R(\hat{y}_{1:t})
    + R(\hat{z}_{1:t}),
\end{equation}
where $R(\hat{y}_{1:t})$ and $R(\hat{z}_{1:t})$ denote the estimated bitrates of the visual latent streams. The weight $\lambda$ controls the balance between rate and distortion. The distortion term $\mathcal{D}$ combines pixel-wise and perceptual losses and is defined as
\begin{equation}
\label{eq:distortion}
    \mathcal{D}
    =
    \| \hat{x}_{1:T} - x_{1:T} \|_1
    +
    \operatorname{LPIPS}(\hat{x}_{1:T}, x_{1:T}).
\end{equation}

\noindent\textbf{Stage~2. Semantic-Guidance Fine-tuning.}
Starting from the Stage~1 checkpoint, we jointly train all Stage~1 parameters, the adapter $\phi$, and the three semantic-guidance modules. The Video-LLM and text encoder remain frozen. Adding the description bitrate and a joint video-text loss gives
\begin{equation}
\label{eq:stage2_objective}
    \mathcal{L}_2=\mathcal{L}_1+R(s_{1:t})+\mathcal{L}_{\mathrm{jvt}},
\end{equation}
where $R(s_{1:t})$ denotes the bitrate of the complete description sequence. Inspired by the joint image-text loss in~\cite{taco_2024}, the joint video-text loss aligns each reconstructed frame group with its corresponding description and is defined as

\begin{equation}
\label{eq:joint_video_semantic_loss}
\begin{aligned}
\mathcal{L}_{\mathrm{jvt}}
={}&
\beta\Biggl[
\frac{1}{t}\sum_{\tau=1}^{t}
\mathcal{L}_{\mathrm{con}}\bigl(
u_\tau(\hat{x}),\, f_T(s_\tau)
\bigr)
\\
&+
\frac{\gamma}{T}\sum_{k=1}^{T}
\left\|
f_I(x_k) - f_I(\hat{x}_k)
\right\|_2\Biggr].
\end{aligned}
\end{equation}
Here, $f_I$ and $f_T$ denote the frozen image and text encoders with normalized outputs. The group feature $u_\tau(\hat{x})$ is the normalized mean of $f_I(\hat{x}_k)$ over $k\in\mathcal{G}_\tau$. The symmetric contrastive loss is computed across the video batch at each $\tau$ and averaged over temporal latents. The second term preserves visual features of individual frames. The overall semantic loss is weighted by $\beta$, and the feature distance term by $\gamma$.

\section{Experiments}
\label{sec:experiments}

\begin{figure*}[p]
  \centering
  \begin{minipage}[c][0.995\textheight][s]{\textwidth}
    \centering
    \includegraphics[width=0.96\linewidth]{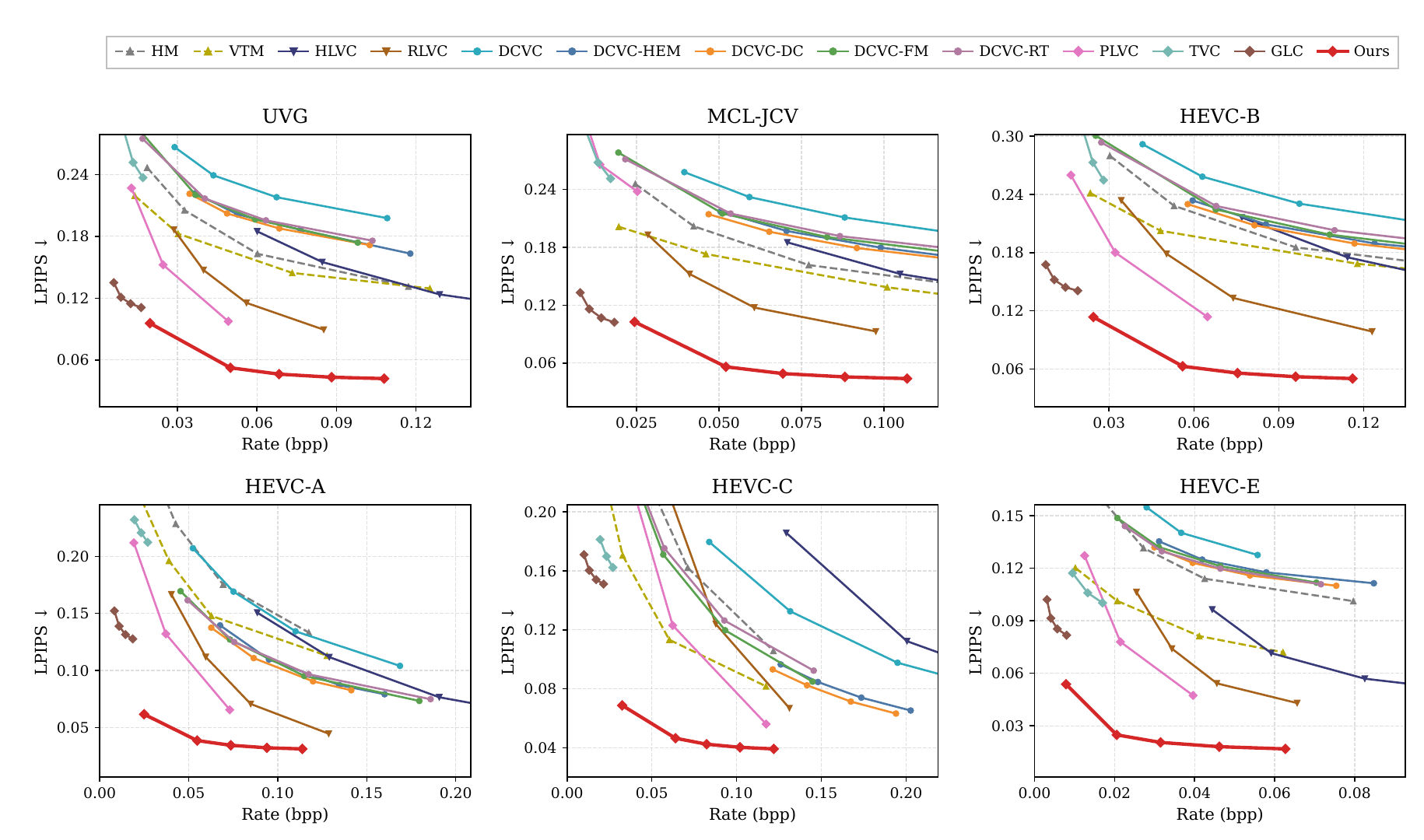}
  \caption{LPIPS rate-distortion curves on UVG, MCL-JCV, HEVC-A, HEVC-B, HEVC-C, and HEVC-E. Lower LPIPS indicates better perceptual quality. Bitrate is measured in bits per pixel (bpp).}
  \label{fig:rdcurves_lpips}
    \vfill
    \includegraphics[width=0.96\linewidth]{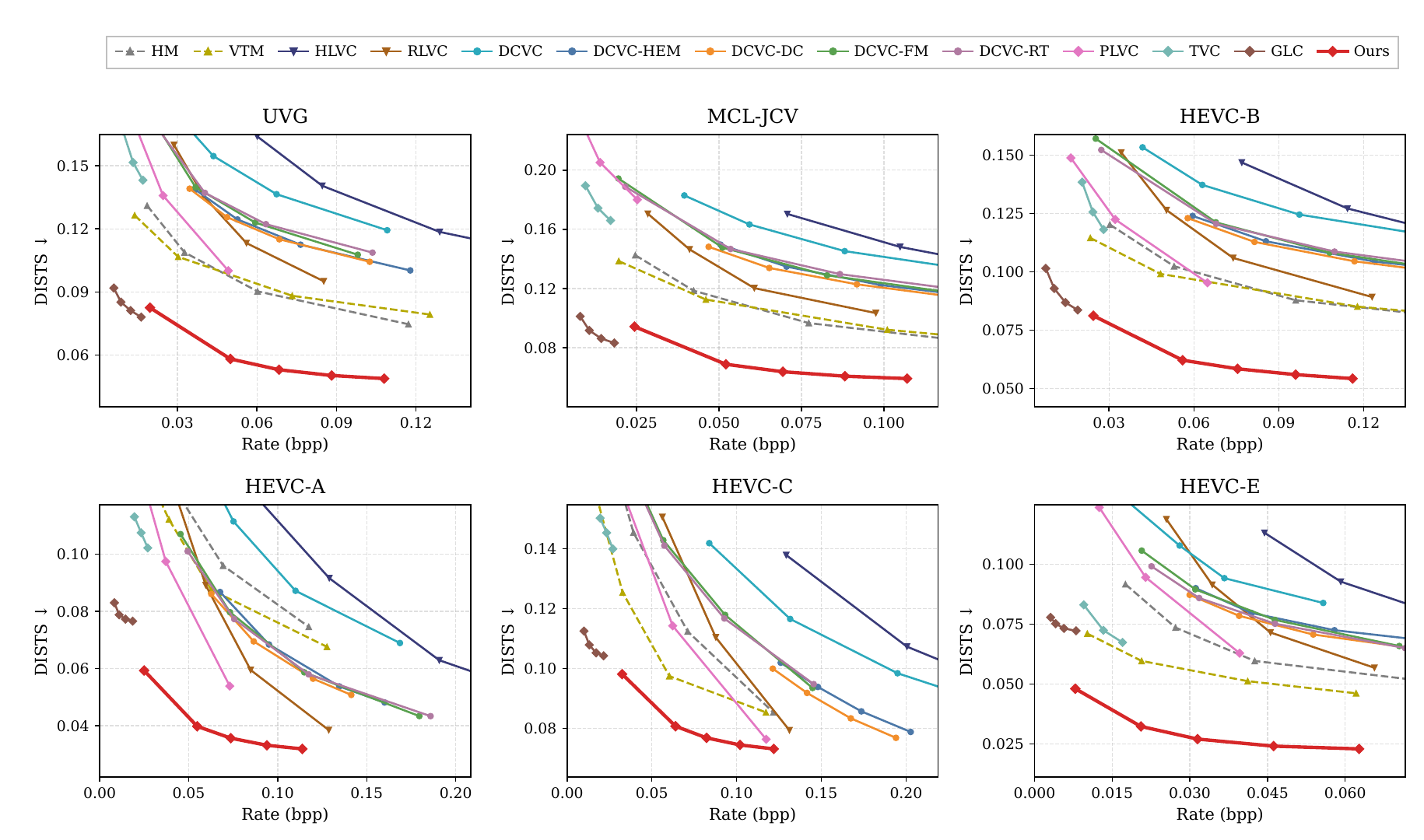}
  \caption{DISTS rate-distortion curves on UVG, MCL-JCV, HEVC-A, HEVC-B, HEVC-C, and HEVC-E. Lower DISTS indicates better perceptual similarity in structure and texture. Bitrate is measured in bits per pixel (bpp).}
  \label{fig:rdcurves_dists}
  \end{minipage}
\end{figure*}

\begin{figure*}[!t]
  \centering
  \includegraphics[width=\linewidth]{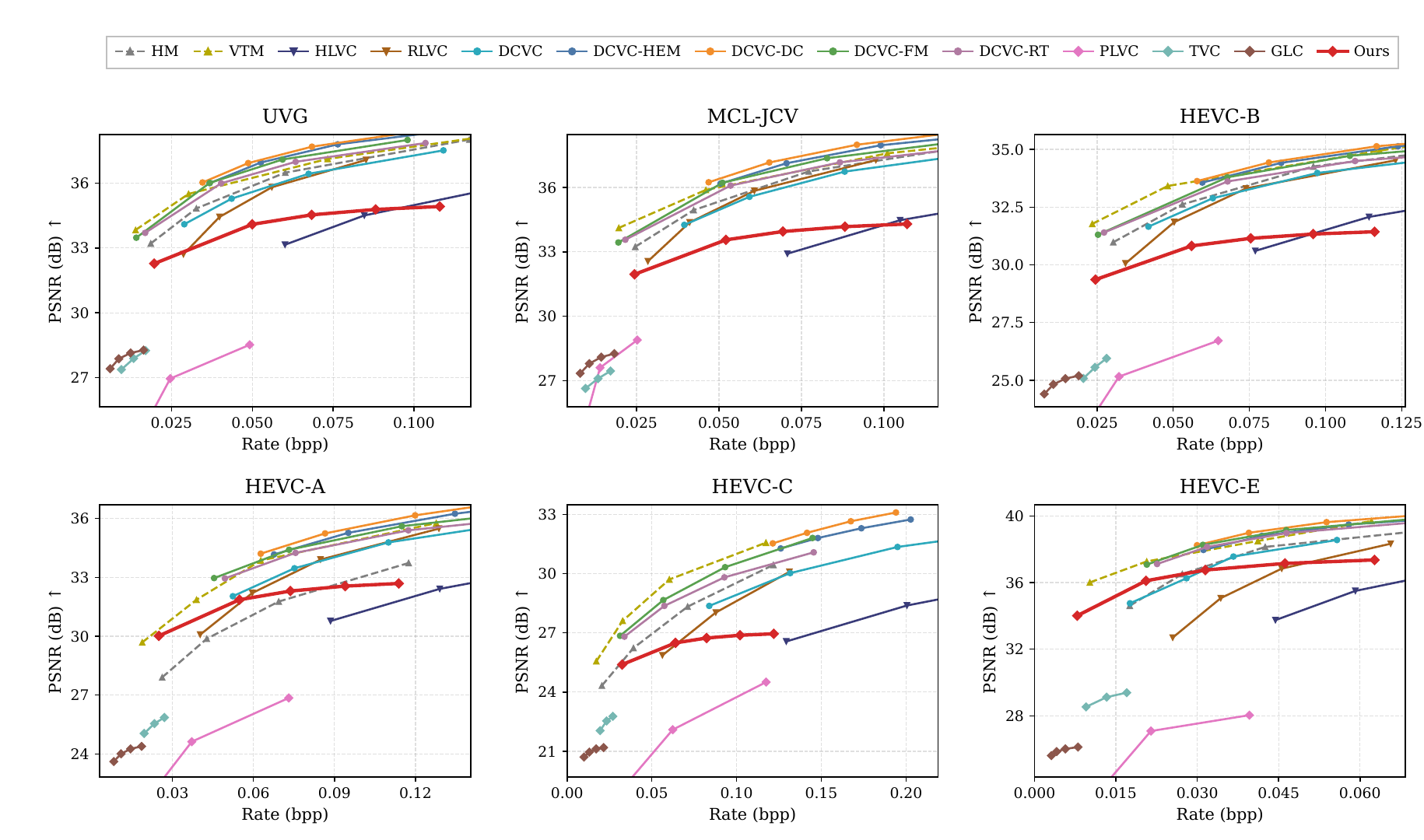}
  \caption{PSNR rate-distortion curves on UVG, MCL-JCV, HEVC-A, HEVC-B, HEVC-C, and HEVC-E. Higher PSNR indicates better reconstruction fidelity. Bitrate is measured in bits per pixel (bpp).}
  \label{fig:rdcurves_psnr}
\end{figure*}

\subsection{Experimental Settings}

\noindent\textbf{Datasets.} We train on Kinetics-600~\cite{k600-02,k600-01}, selecting clips with at least 97 frames. We use the first 97 frames of each clip and apply a $256 \times 256$ center crop. We evaluate on UVG~\cite{uvg}, MCL-JCV~\cite{mcl-jcv}, HEVC-A, HEVC-B, HEVC-C, and HEVC-E~\cite{hevc}. These benchmarks cover diverse content types and spatial resolutions.

\noindent\textbf{Implementation Details.} We use LeanVAE~\cite{leanvae} with $4{\times}8{\times}8$ spatiotemporal downsampling ($r{=}4$) as the video tokenizer, producing latents $l_{1:t} \in \mathbb{R}^{(1+(T-1)/4)\times H/8 \times W/8 \times C}$ with $C{=}16$. We use Qwen3-VL-2B-Instruct~\cite{qwen3vl2b} to generate one description for the first frame and one for each subsequent group of four frames, matching the temporal latent slices. The resulting descriptions are encoded by the frozen Long-CLIP text encoder~\cite{longclip}. We train models with $\lambda \in \{64,256,512,1024,2048\}\times10^{-2}$ to cover a broad bitrate range and set $\beta{=}0.005$ and $\gamma{=}40$ in $\mathcal{L}_{\mathrm{jvt}}$.

\noindent\textbf{Baselines.} Traditional baselines include HM 16.25~\cite{hm} and VTM 17.0~\cite{vtm} under the Low Delay configuration, while neural baselines include HLVC~\cite{hlvc}, RLVC~\cite{rlvc}, DCVC~\cite{dcvc}, DCVC-HEM~\cite{dcvc-hem}, DCVC-DC~\cite{dcvc-dc}, DCVC-FM~\cite{dcvc-fm}, DCVC-RT~\cite{dcvc-rt}, and perceptual codecs PLVC~\cite{plvc}, TVC~\cite{tvc}, and GLC~\cite{glc}. All methods are evaluated on the first 96 frames, with an intra period of 96 for the baselines. LLMVC repeats the last frame to form a 97-frame input sequence for tokenization, while reconstruction quality is evaluated on the original 96 frames.

\noindent\textbf{Metrics.} We evaluate perceptual quality using LPIPS~\cite{lpips} and DISTS~\cite{dists}, where lower values indicate better quality. We also report PSNR as a complementary measure of reconstruction fidelity, where higher values indicate better quality.

\subsection{Quantitative Performance}
Figures~\ref{fig:rdcurves_lpips} and~\ref{fig:rdcurves_dists} compare perceptual rate-distortion performance across six benchmarks. LLMVC consistently achieves lower LPIPS and DISTS than traditional codecs and neural codecs optimized for fidelity. Compared with PLVC, LLMVC consistently achieves better performance. For example, LLMVC achieves LPIPS and DISTS BD-rate savings of 67.1\% and 50.4\% on HEVC-C. On UVG, MCL-JCV, and HEVC-B, LLMVC provides perceptual quality comparable to GLC at LLMVC's lowest tested rates, with further improvements as bitrate increases. On HEVC-A, HEVC-C, and HEVC-E, LLMVC achieves lower LPIPS and DISTS than GLC. These results support the perceptual coding benefits of the semantic-guidance modules.

Figure~\ref{fig:rdcurves_psnr} shows that LLMVC achieves substantially higher PSNR than perceptual codecs such as PLVC and GLC across all six datasets, while approaching the fidelity of distortion-oriented codecs on HEVC-E.

\subsection{Qualitative Performance}

\begin{figure*}[!t]
  \centering
  \includegraphics[width=1.0\linewidth]{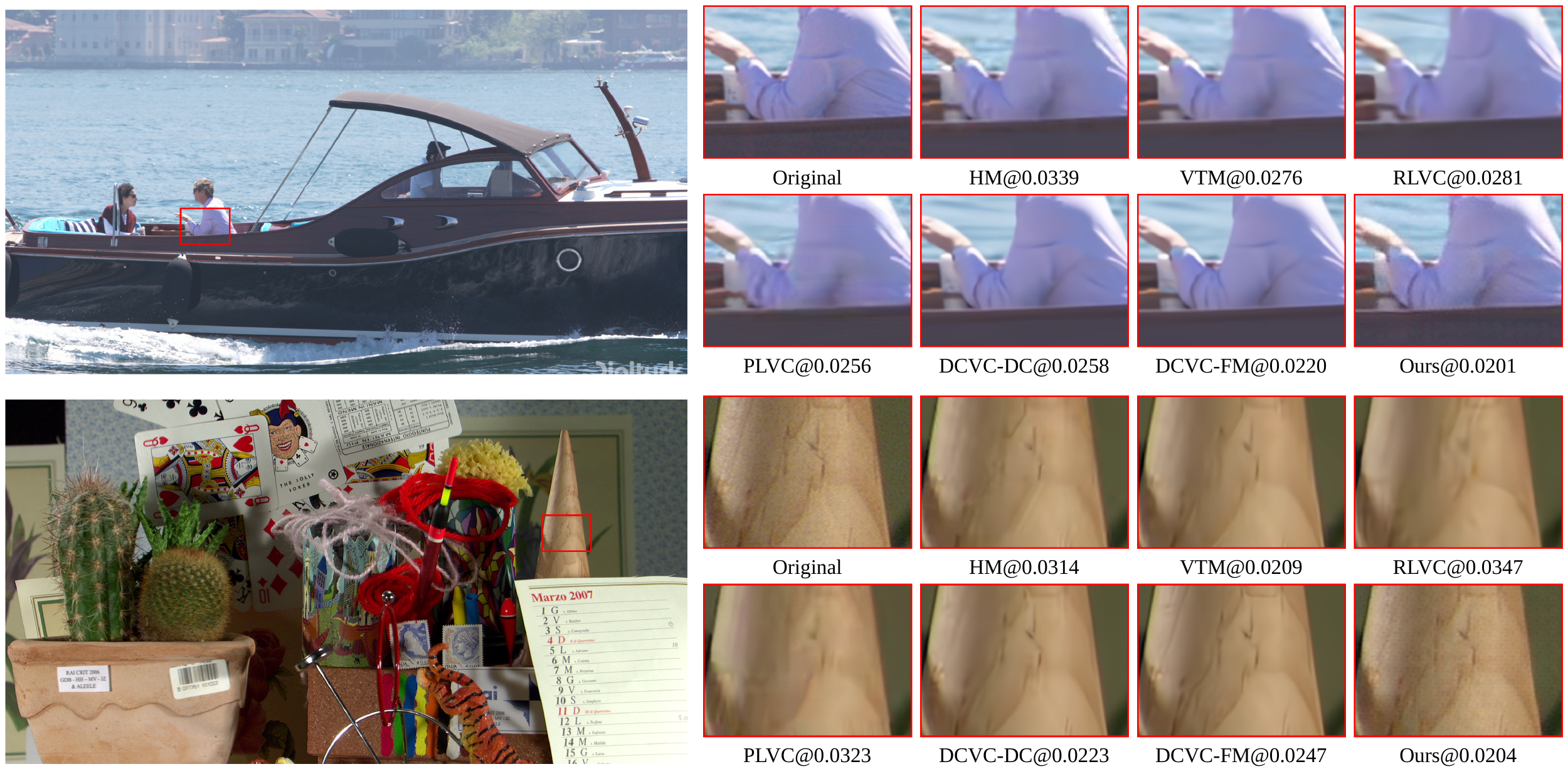}
  \caption{
    Qualitative comparisons on YachtRide (UVG) and Cactus (HEVC-B). Enlarged views show the regions marked in red. Labels follow the format Method@bpp, where bpp denotes bits per pixel.
  }
  \label{fig:visual_results}
\end{figure*}

\begin{table}[!t]
\centering
\caption{Bitrate decomposition on UVG in bits per pixel (bpp). Subscripts $y$, $z$, and $s$ denote the compact representation, hyperlatent, and semantic descriptions. The $\mathrm{bpp}_s$ ratio denotes the percentage of the total bitrate used for semantic descriptions.}
\label{tab:bpp_decomp}
\footnotesize
\setlength{\tabcolsep}{6pt}
\renewcommand{\arraystretch}{1.15}
\begin{tabular*}{\columnwidth}{@{\extracolsep{\fill}}cccccc@{}}
\toprule
$\lambda\;(\times 10^{-2})$ & bpp & bpp$_y$ & bpp$_z$ & bpp$_s$ & bpp$_s$ ratio \\
\midrule
64 & 0.0338 & 0.0240 & 0.0094 & 0.0004 & 1.15\% \\
512 & 0.0948 & 0.0816 & 0.0128 & 0.0004 & 0.41\% \\
2048 & 0.1459 & 0.1280 & 0.0175 & 0.0004 & 0.27\% \\
\bottomrule
\end{tabular*}
\end{table}

Figure~\ref{fig:visual_results} presents qualitative comparisons on YachtRide from UVG and Cactus from HEVC-B. On YachtRide, LLMVC preserves visible clothing texture and folds that are smoothed in several competing reconstructions. On Cactus, it retains the fine surface texture in the enlarged region, producing a more detailed appearance closer to the original frame. These details are preserved at the lowest bitrate among the compared methods in both examples, demonstrating the effectiveness of LLMVC in maintaining perceptually important textures under limited bitrate budgets. Frame-wise generative codecs such as GLC may also exhibit temporal flickering, with textures changing inconsistently across frames.

\subsection{Ablation Study}

\begin{table}[!t]
\centering
\caption{Semantic-guidance ablation on HEVC-C. $\surd$ and $\times$ indicate enabled and disabled modules. Negative LPIPS BD-rate values indicate average bitrate savings relative to the baseline at matched LPIPS.}
\label{tab:ablation_modules}
\footnotesize
\setlength{\tabcolsep}{4.5pt}
\renewcommand{\arraystretch}{1.25}
\begin{tabular*}{\columnwidth}{@{\extracolsep{\fill}}lcccc@{}}
\toprule
Configuration & SLA & SCEM & STE & LPIPS BD-rate $\downarrow$ \\
\midrule
Baseline & $\times$ & $\times$ & $\times$ & \textemdash \\
\midrule
\quad with SLA & $\surd$ & $\times$ & $\times$ & $-4.84\%$ \\
\quad with SLA+SCEM & $\surd$ & $\surd$ & $\times$ & $-6.00\%$ \\
\quad with SLA+SCEM+STE & $\surd$ & $\surd$ & $\surd$ & $\mathbf{-7.92\%}$ \\
\bottomrule
\end{tabular*}
\end{table}

\noindent\textbf{Bitrate Decomposition.}
Table~\ref{tab:bpp_decomp} shows that reusing the same descriptions across quality levels keeps the semantic bitrate at approximately 0.0004 bpp. Its share decreases from 1.15\% to 0.27\% as total bitrate increases, supporting the semantic-guidance modules with little transmission overhead.

\noindent\textbf{Semantic Module Ablation.}
Table~\ref{tab:ablation_modules} evaluates the three semantic-guidance modules on HEVC-C using LPIPS BD-rate relative to the baseline without these modules. Adding SLA yields a bitrate saving of 4.84\%. Incorporating SCEM increases the saving to 6.00\%, and further adding STE achieves the best result of 7.92\%. These progressive gains support the complementary roles of SLA, SCEM, and STE across the compression pipeline.

\subsection{Runtime Analysis}

LLMVC achieves the fastest decoding among the compared methods at 191 ms/frame for 1080p videos (Table~\ref{tab:complexity}). It is 2 times as fast as DCVC-DC and DCVC-FM and 1.5 times as fast as GLC. Encoding takes 341 ms/frame including Video-LLM inference and 229 ms/frame without it, both faster than DCVC-DC and DCVC-FM. The decoder uses transmitted descriptions without Video-LLM inference.

\begin{table}[t]
\centering
\caption{Average encoding and decoding time at 1080p resolution on an AMD EPYC 7513 CPU and a single NVIDIA L40S GPU, measured in ms/frame. Bold and underlined values indicate the fastest and second fastest methods in each column.}
\label{tab:complexity}
\footnotesize
\setlength{\tabcolsep}{6pt}
\begin{tabular}{lcc}
\toprule
Method & Enc. (ms/frame) & Dec. (ms/frame) \\
\midrule
DCVC-DC  & 502 & 383 \\
DCVC-FM  & 455 & 388 \\
GLC      & \textbf{192} & \underline{280} \\
\midrule
\textbf{LLMVC (Ours)} & \underline{341} & \textbf{191} \\
\bottomrule
\end{tabular}
\end{table}

\section{Conclusion}
\label{sec:conclusion}
We presented LLMVC, an LLM-guided generative video compression framework that introduces a compact, temporally aligned semantic channel to guide latent modeling throughout the codec. HBBA enables SLA and STE to provide semantic alignment at the encoder and semantic refinement at the decoder, while SCCA supports probability conditioning in SCEM. Experiments on six benchmarks show consistent improvements in LPIPS and DISTS, demonstrating that language-derived semantic priors can effectively guide visual latents for perceptually faithful reconstruction.

\noindent\textbf{Limitations and Future Work.}
The frozen Video-LLM bounds semantic quality, limiting guidance for complex or rare scenes. At ultra-low bitrates, semantic features provide high-level context but cannot recover fine details lost to heavy quantization, leading to diminishing gains. Future work will explore adapting the Video-LLM for compression-aware captioning and generative decoding strategies that leverage semantic priors to synthesize lost details at extreme compression ratios.

{
    \small
    \bibliographystyle{ieeenat_fullname_citation}
    \bibliography{main}
}

\clearpage
\setcounter{section}{0}
\setcounter{subsection}{0}
\setcounter{figure}{0}
\setcounter{table}{0}
\setcounter{equation}{0}
\renewcommand{\thesection}{S\arabic{section}}
\renewcommand{\thesubsection}{\thesection.\arabic{subsection}}
\renewcommand{\thefigure}{S\arabic{figure}}
\renewcommand{\thetable}{S\arabic{table}}
\renewcommand{\theequation}{S\arabic{equation}}
\setkeys{Gin}{height=0.85\textheight,keepaspectratio}
\pdfbookmark[0]{Supplementary Material}{supplementary-material}
\maketitlesupplementary

\begin{figure*}[!htbp]
  \centering
  \includegraphics[width=1.0\linewidth]{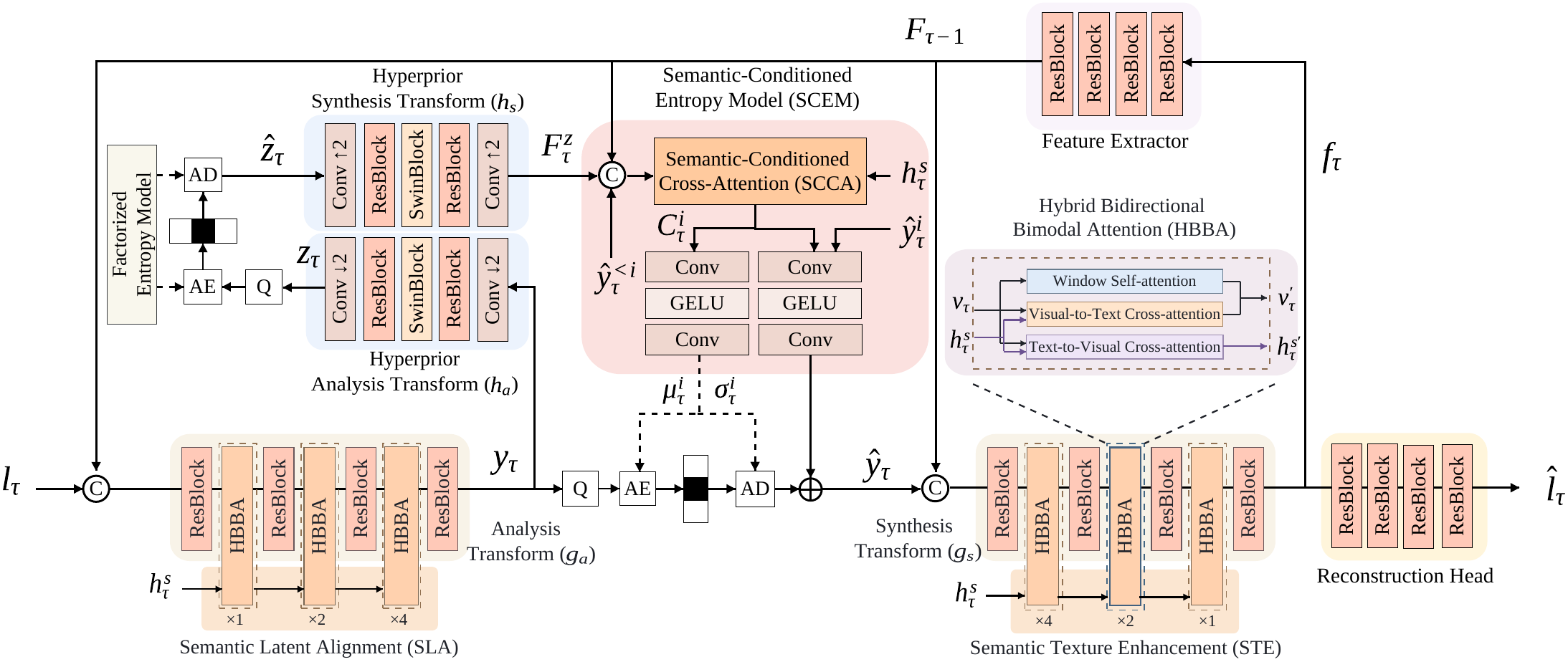}
  \caption{Network architecture of the LLMVC latent codec with semantic-guidance modules. The analysis transform $g_a$ combines $l_\tau$ with the previous temporal context $F_{\tau-1}$ and alternates residual blocks with Semantic Latent Alignment (SLA) interactions to produce $y_\tau$. The hyperprior transforms $h_a$ and $h_s$ produce the hyperlatent $z_\tau$ and recover features $F_\tau^z$ from $\hat{z}_\tau$. The Semantic-Conditioned Entropy Model (SCEM) uses $F_\tau^z$, $F_{\tau-1}$, previously decoded slices $\hat{y}_\tau^{<i}$, and semantic features $h_\tau^s$ to form context $C_\tau^i$ for predicting the Gaussian mean $\mu_\tau^i$ and scale $\sigma_\tau^i$. The synthesis transform $g_s$ processes $\hat{y}_\tau$ and $F_{\tau-1}$ with alternating residual blocks and Semantic Texture Enhancement (STE) interactions. Its output $f_\tau$ is used to reconstruct $\hat{l}_\tau$ and generate the next temporal context $F_\tau$. SLA and STE use Hybrid Bidirectional Bimodal Attention (HBBA) and pass updated semantic features along their own paths. Q denotes quantization, AE and AD denote arithmetic encoding and decoding, and \textcircled{\scalebox{0.68}{C}} denotes concatenation.}
  \label{fig:arch_detail}
\end{figure*}

\section{HM and VTM Test Settings}
\label{sec:traditional_codec}

We evaluate HM-16.25 and VTM-17.0 on UVG, MCL-JCV, HEVC-A, HEVC-B, HEVC-C, and HEVC-E. For each sequence, we encode the first 96 frames using official low-delay configurations, with codec settings following those adopted in DCVC-DC and DCVC-FM. Different rate points are obtained by varying the quantization parameter (QP) while keeping all other coding parameters fixed. Before encoding, the 8-bit YUV420 source sequences are converted to 10-bit YUV444 through chroma upsampling and bit-depth conversion. The original and reconstructed YUV444 sequences are then converted to RGB using BT.709, and evaluation metrics are computed in the RGB domain. Bitrate is measured from the actual coded bitstream.

For HM-16.25, we use \texttt{TAppEncoderStatic} with the official
\texttt{encoder\_lowdelay\_main\_rext.cfg} configuration. For
VTM-17.0, we use \texttt{EncoderAppStatic} with the official
\texttt{encoder\_lowdelay\_vtm.cfg} configuration. Both codecs use the following  settings:

\begin{quote}
\small\ttfamily
\hspace*{1em}-c \{\textnormal{\textit{config file name}}\}\\
\hspace*{1em}-{}-InputFile=\{\textnormal{\textit{input file name}}\}\\
\hspace*{1em}-{}-InputBitDepth=10\\
\hspace*{1em}-{}-OutputBitDepth=10\\
\hspace*{1em}-{}-OutputBitDepthC=10\\
\hspace*{1em}-{}-InputChromaFormat=444\\
\hspace*{1em}-{}-FrameRate=\{\textnormal{\textit{frame rate}}\}\\
\hspace*{1em}-{}-DecodingRefreshType=2\\
\hspace*{1em}-{}-FramesToBeEncoded=\{\textnormal{\textit{frame number}}\}\\
\hspace*{1em}-{}-SourceWidth=\{\textnormal{\textit{width}}\}\\
\hspace*{1em}-{}-SourceHeight=\{\textnormal{\textit{height}}\}\\
\hspace*{1em}-{}-IntraPeriod=96\\
\hspace*{1em}-{}-QP=\{\textnormal{\textit{qp}}\}\\
\hspace*{1em}-{}-Level=6.2\\
\hspace*{1em}-{}-BitstreamFile=\{\textnormal{\textit{bitstream file name}}\}
\end{quote}

\section{Network Architecture Details}
\label{sec:arch}

Figure~\ref{fig:arch_detail} shows the network architecture of the LLMVC latent codec. For each temporal latent $l_\tau$, the codec uses semantic features $h_\tau^s$ derived from the corresponding description $s_\tau$. The analysis transform $g_a$ combines the temporal latent $l_\tau$ with the context $F_{\tau-1}$ to produce the compact representation $y_\tau$. Its residual blocks alternate with SLA interactions that align intermediate visual features with semantic features through HBBA. The synthesis transform $g_s$ follows the same design with STE. SLA and STE each receive $h_\tau^s$ as their initial semantic features and pass the updated features between successive HBBA interactions.

The hyperprior analysis transform $h_a$ and synthesis transform $h_s$ both use residual and Swin blocks. The transform $h_a$ uses strided convolutions to downsample $y_\tau$ by a factor of four along each spatial dimension, producing the hyperlatent $z_\tau$. The transform $h_s$ uses convolutional upsampling to recover the hyperprior features $F_\tau^z$ from the decoded hyperlatent $\hat{z}_\tau$. The codec encodes $y_\tau$ in four channel slices. For slice $i$, SCEM uses SCCA to combine $h_\tau^s$ with the visual context formed by $F_\tau^z$, $F_{\tau-1}$, and all previously decoded slices $\hat{y}_\tau^{<i}$. The resulting context $C_\tau^i$ guides prediction of the Gaussian mean $\mu_\tau^i$ and scale $\sigma_\tau^i$ for entropy coding.

The decoded slices are concatenated into $\hat{y}_\tau$, which enters the synthesis path with $F_{\tau-1}$. From the resulting features $f_\tau$, the reconstruction head produces $\hat{l}_\tau$ and the feature extractor generates $F_\tau$ for the next latent.

The representations $y_\tau$ and $z_\tau$ have 256 and 128 channels, respectively. Window self-attention in HBBA uses 8, 9, and 6 heads for visual feature dimensions of 64, 144, and 192 channels, with corresponding dimensions of 8, 16, and 32 channels per head. These configurations are used in this order along $g_a$ and in reverse order along $g_s$. Visual-to-text and text-to-visual cross-attention use the same head configurations at the corresponding feature dimensions. Window self-attention uses regular and shifted $8\times8$ windows. The hyperprior Swin blocks use $4\times4$ windows. The cross-attention in SCCA uses 16 heads with 32 channels per head. A trainable two-layer MLP adapter maps Long-CLIP features to semantic features.

\section{Semantic Description Generation and Transmission}
\label{sec:description_generation}

We expand on the semantic description generation and coding introduced in Step~1 of Section~3.3.

\noindent\textbf{Semantic Description Generation.} We use frozen Qwen3-VL-2B-Instruct to describe frame groups $\{x_1\}$, $\{x_2,\ldots,x_5\}$, $\{x_6,\ldots,x_9\}$, $\ldots$, following LeanVAE's temporal grouping. Each group is provided in temporal order to generate a description $s_\tau$ aligned with temporal latent $l_\tau$.

We use separate prompts for the first frame and subsequent four-frame groups. The first-frame prompt captures object appearance, background, and spatial layout. The four-frame prompt summarizes each subsequent group as a whole, capturing object appearance, motion, and camera movement.

\noindent{\small\textbf{\textit{First-frame prompt.}} \textit{``Describe this video frame to help a video compression model. Cover the main objects with their colors and textures, the background, and the spatial layout. Only describe what is clearly visible.''}}

\begin{samepage}
\noindent{\small\textbf{\textit{Four-frame prompt.}} \textit{``These are 4 consecutive frames of one video segment shown in temporal order. Describe the segment as a whole to help a video compression model. Cover the main objects with their appearance, how they move and change during the segment, and any camera movement. Do not describe each frame separately. Only describe what is clearly visible.''}}
\par
\end{samepage}

\begin{figure*}[p]
  \centering
  \begin{minipage}{\textwidth}
    \centering
    \includegraphics[width=\linewidth]{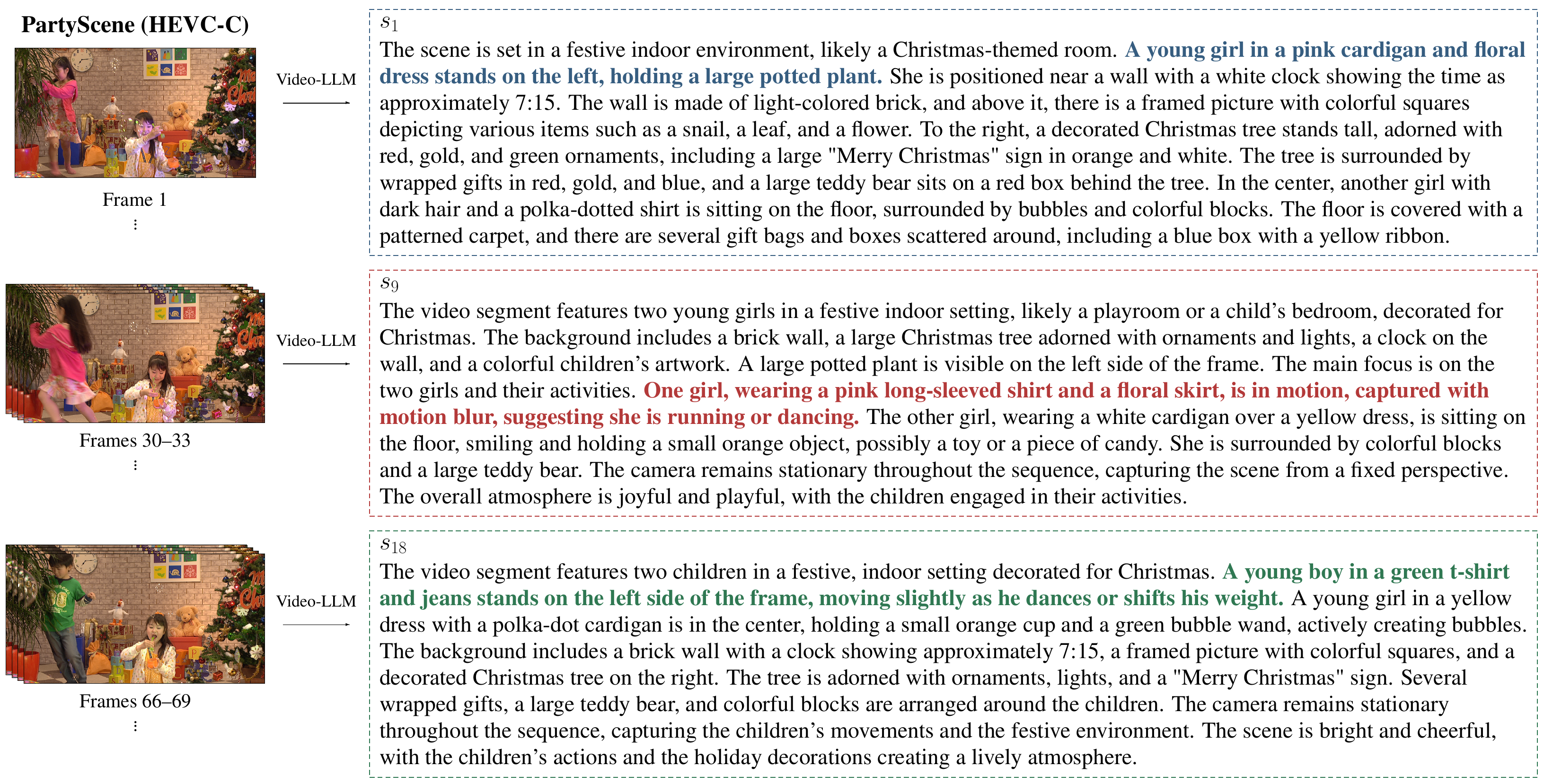}
  \caption{Examples of descriptions aligned with temporal latents on PartyScene (HEVC-C). Descriptions $s_1$, $s_9$, and $s_{18}$ correspond to frame 1, frames 30--33, and frames 66--69, respectively. Colored bold sentences highlight differences in descriptive content across frame groups.}
  \label{fig:description_generation}
    \par\vspace{12pt}
    \includegraphics[width=0.96\linewidth]{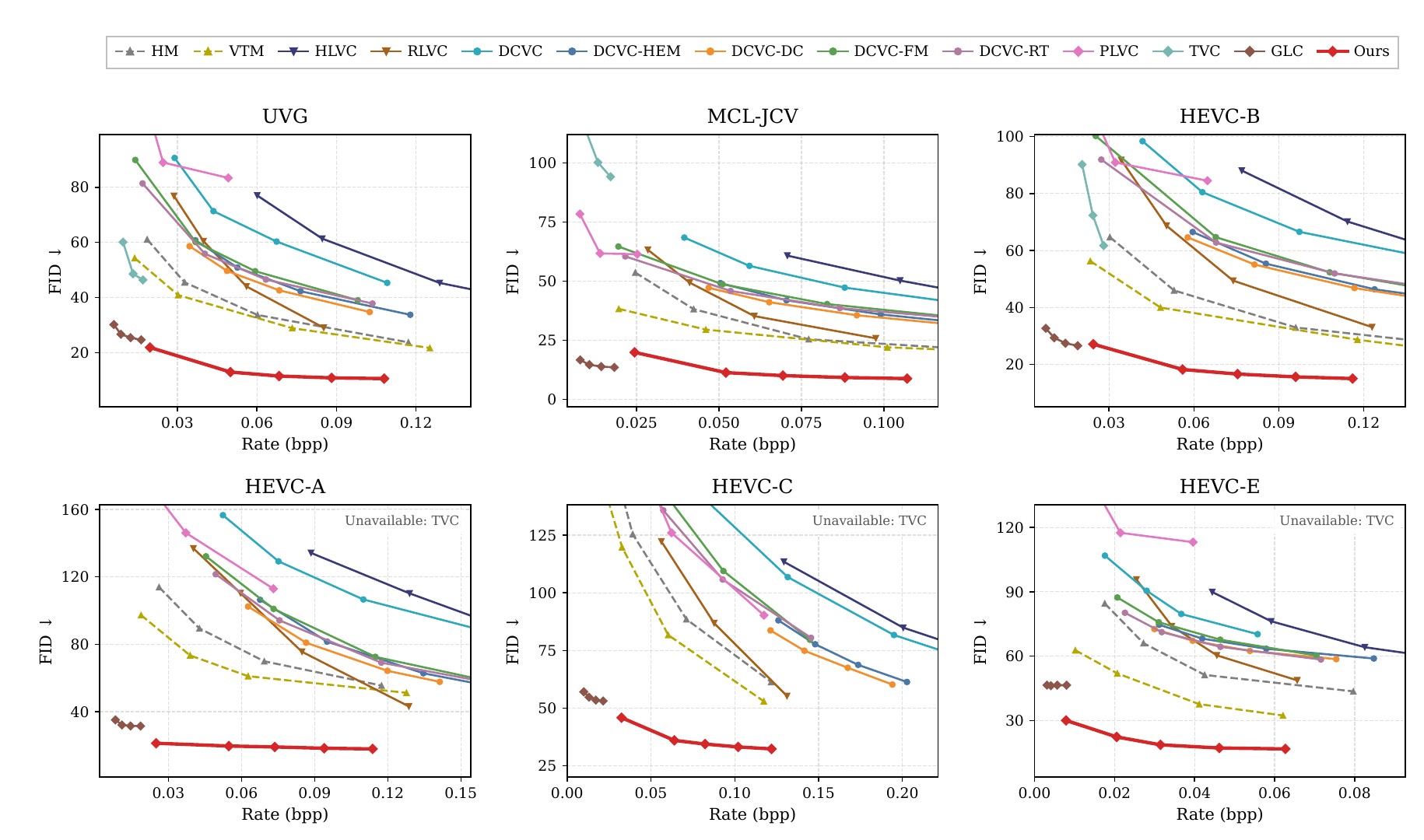}
    \caption{FID rate-distortion curves on UVG, MCL-JCV, HEVC-A, HEVC-B, HEVC-C, and HEVC-E. Lower FID indicates greater similarity between the feature distributions of original and reconstructed videos. Bitrate is measured in bits per pixel (bpp).}
    \label{fig:fid_rd}
  \end{minipage}
\end{figure*}

Figure~\ref{fig:description_generation} illustrates how descriptions remain aligned with the video content over time in PartyScene from HEVC-C. The first description $s_1$ captures the girl in pink holding a potted plant. As she begins running, $s_9$ updates the description to reflect her movement. When a boy in green enters the scene, $s_{18}$ captures the new subject and his movement. These updates align the semantic information with the corresponding temporal latents, providing semantic features consistent with the changing visual content.

We use different procedures to generate descriptions for training and testing. We first generate and store descriptions for all videos in the training set. During codec training, the stored descriptions are loaded with their corresponding video clips, avoiding repeated Video-LLM inference. At test time, the encoder runs the Video-LLM on each input video to generate descriptions online.

\noindent\textbf{Semantic Description Transmission.} Each description $s_\tau$ is converted into token IDs using the Long-CLIP tokenizer. These IDs are serialized into byte symbols and losslessly compressed using adaptive arithmetic coding. Symbol probabilities are initialized uniformly and estimated from cumulative occurrence counts, which are updated after each symbol. The decoder reproduces these updates to recover the token IDs, while stored description lengths preserve their alignment with temporal latents. Both ends construct identical 248-token Long-CLIP input windows from these IDs. For longer descriptions, features from multiple windows are concatenated at the corresponding temporal latent index, with padding masked. The frozen text encoder and trainable adapter produce the semantic features $h_\tau^s$. The semantic bitrate includes the complete description sequence.

\section{More Optimization Details}
\label{sec:more_implementation_details}

We follow the two-stage optimization described in Section~3.4. We use AdamW with $\beta_1=0.9$, $\beta_2=0.95$, and weight decay $10^{-4}$. A separate Adam optimizer updates the entropy-bottleneck quantile parameters with an initial learning rate of $10^{-3}$. Training uses gradient clipping at norm 1.0 and BF16 mixed precision on a single machine with eight NVIDIA L40S GPUs.

Specifically, in Stage~1, the tokenizer and latent codec use cosine learning rate decay from $10^{-4}$ to $10^{-5}$. We progressively increase the clip length through $T\in\{9,17,33,49,97\}$, training for 20 epochs at each length with corresponding per-GPU batch sizes of $\{16,16,16,8,8\}$. Stage~2 initializes from the Stage~1 checkpoint, with cosine learning rate decay from $10^{-5}$ to $10^{-6}$ for the tokenizer and latent codec and from $10^{-4}$ to $10^{-5}$ for the adapter and semantic-guidance modules. We use the same clip lengths and batch sizes, training for 10 epochs at each length.

\section{More Complexity Analysis}
\label{sec:more_complexity}

\suppressfloats[t]
\begin{table}[t]
\centering
\caption{Parameter counts and estimated FLOPs on UVG at 1080p resolution. \#Params (M) denotes millions of parameters, and FLOPs (T) denotes trillions of floating-point operations per frame. FLOPs cover the 97-frame input for LLMVC and are averaged over the 96 reported frames.}
\label{tab:params}
\footnotesize
\setlength{\tabcolsep}{4pt}
\renewcommand{\arraystretch}{1.2}
\begin{tabular}{@{}lrr@{}}
\toprule
Method & \#Params (M) & FLOPs (T) \\
\midrule
DCVC-DC                    & 50.78   & 5.22 \\
DCVC-FM                    & 44.92   & 4.41 \\
GLC                        & 234.18  & 12.45 \\
\midrule[0.8pt]
LLMVC                      & 2358.72 & 10.62 \\
LLMVC w/o Video-LLM         & 231.18  & 2.37 \\
\bottomrule
\end{tabular}
\end{table}

Complementing the runtime analysis in Section~4.5, Table~\ref{tab:params} compares parameter counts and estimated FLOPs. The complete system contains 2358.72\,M parameters and consists of frozen and trainable components. The frozen components comprise two pretrained semantic models. The Long-CLIP text encoder (63.38\,M) provides semantic features. We use Qwen3-VL-2B-Instruct (2127.53\,M) as an off-the-shelf Video-LLM for description generation without optimizing its parameters. The trainable components comprise the LeanVAE tokenizer (39.79\,M), latent codec (85.18\,M), and semantic-guidance modules (42.84\,M). These components account for only 167.80\,M trainable parameters, representing 7.1\% of the total. This design enables the semantic-guidance modules to integrate pretrained semantic knowledge throughout compression.

As shown in Table~\ref{tab:params}, the complete LLMVC system requires 10.62\,T FLOPs per frame. Excluding Video-LLM inference, the compression pipeline requires 2.37\,T FLOPs per frame, accounting for only 22.3\% of the total. This computational cost is lower than that of DCVC-DC, DCVC-FM, and GLC. Compared with DCVC-DC and DCVC-FM, this represents reductions of 54.6\% and 46.3\%. The computation for description generation is confined to the encoder. The decoder obtains semantic features from the transmitted descriptions without running the Video-LLM, allowing the semantic-guidance modules to support reconstruction while maintaining efficient decoding.

\section{More Quantitative Results}
\label{sec:additional_rd}

\noindent\textbf{Fr\'echet Inception Distance (FID).} Complementing the LPIPS and DISTS results in Section~4.2, Figure~\ref{fig:fid_rd} evaluates perceptual quality using FID. We compute FID using Inception-v3 features of $256\times256$ patches extracted from original and reconstructed frames with a stride of 128 pixels. LLMVC consistently achieves lower FID than codecs optimized for fidelity. Compared with GLC, LLMVC provides comparable performance on UVG, MCL-JCV, and HEVC-B, and achieves lower FID on HEVC-A, HEVC-C, and HEVC-E. These results further support the perceptual coding benefits of the semantic-guidance modules across diverse video content.

\noindent\textbf{PSNR.} Section~4.2 compares the PSNR rate-distortion performance of LLMVC with traditional and neural codecs. Although perceptually trained backbones typically have limited PSNR performance, LLMVC already achieves substantially higher PSNR than perceptual codecs such as PLVC and GLC across all six datasets.

\begin{figure*}[!tp]
  \centering
  \includegraphics[width=0.96\linewidth]{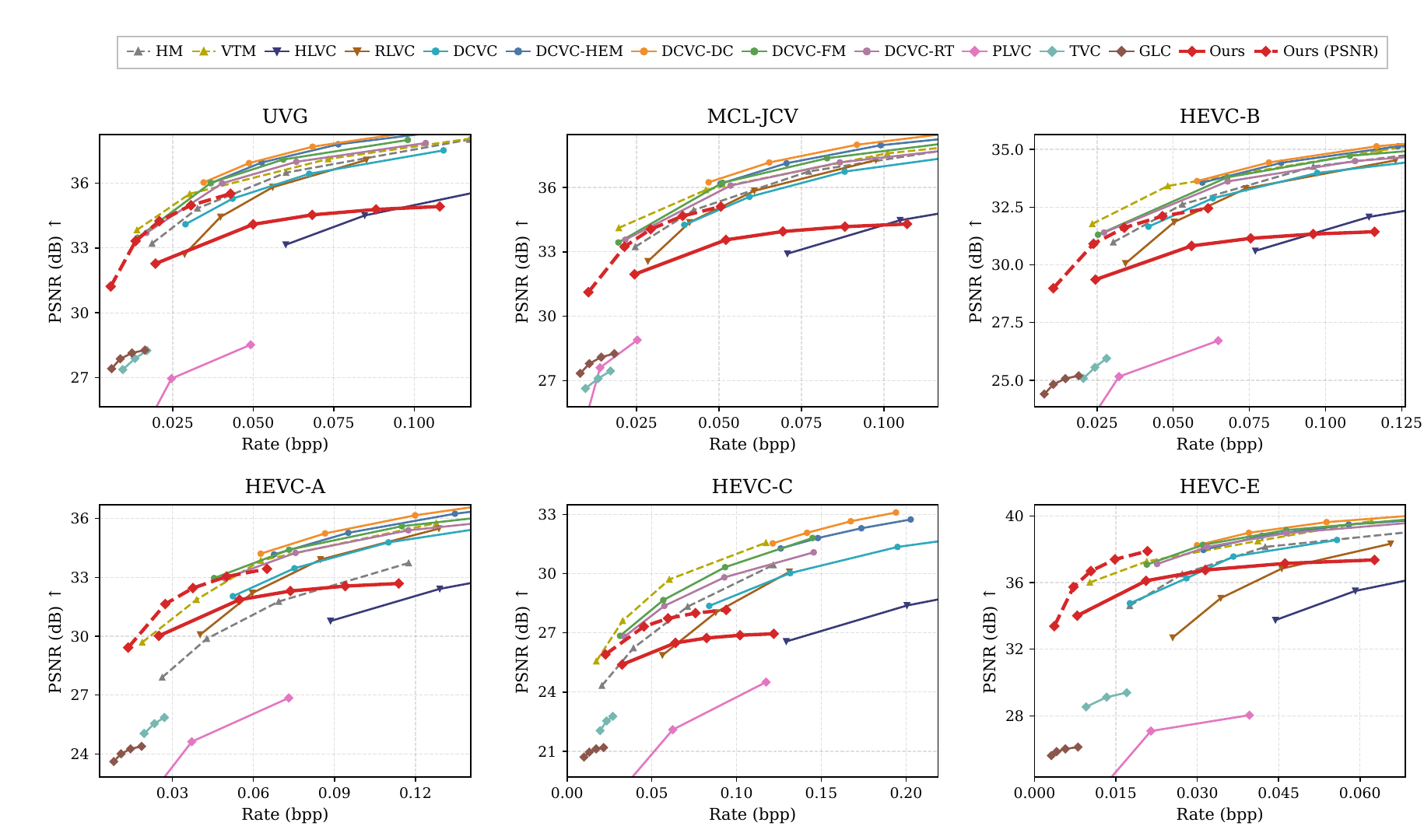}
    \caption{PSNR rate-distortion curves on UVG, MCL-JCV, HEVC-A, HEVC-B, HEVC-C, and HEVC-E. Ours (PSNR) denotes LLMVC fine-tuned using MSE as the distortion term. Higher PSNR indicates better reconstruction fidelity. Bitrate is measured in bits per pixel (bpp).}
    \label{fig:psnr_finetuning}
\end{figure*}

To evaluate the flexibility of LLMVC under different distortion objectives, we fine-tune LLMVC using mean squared error (MSE) as the distortion term in the compression objective, yielding LLMVC (PSNR). Figure~\ref{fig:psnr_finetuning} compares its PSNR rate-distortion performance across six datasets. At low bitrates, LLMVC (PSNR) achieves performance comparable to codecs optimized for fidelity, with particularly strong results on HEVC-A and HEVC-E.

\section{More Qualitative Results}

\begin{figure*}[!htbp]
  \centering
  \includegraphics[width=\linewidth,height=0.94\textheight,keepaspectratio]{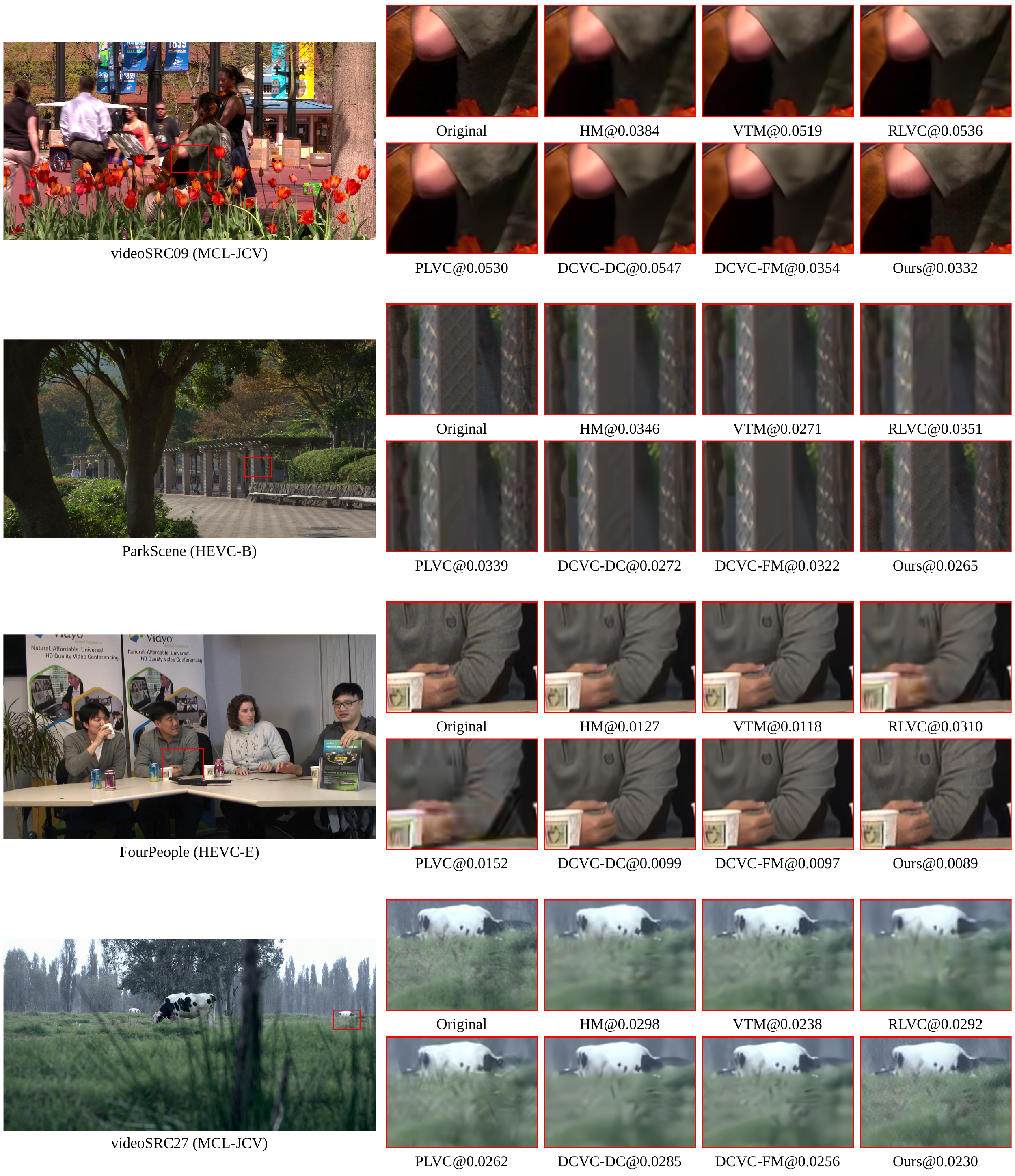}
  \caption{Qualitative comparisons on videoSRC09 (MCL-JCV), ParkScene (HEVC-B), FourPeople (HEVC-E), and videoSRC27 (MCL-JCV), from top to bottom. Enlarged views show the regions marked in red. Labels follow the format Method@bpp, where bpp denotes bits per pixel.}
  \label{fig:visual_more}
\end{figure*}

Figure~\ref{fig:visual_more} presents qualitative comparisons across four video sequences. On videoSRC09 (MCL-JCV) and FourPeople (HEVC-E), LLMVC retains clothing texture that is smoothed in several competing reconstructions. It also preserves the repeated pillar patterns in ParkScene (HEVC-B) and more visible grass texture around the distant cow in videoSRC27 (MCL-JCV). These details are retained at the lowest bitrate among the compared methods in all four examples, demonstrating the effectiveness of LLMVC in preserving perceptually important textures under limited bitrate budgets.

\end{document}